\documentclass[10pt,twocolumn,letterpaper]{article}
\usepackage[accsupp]{axessibility}
\usepackage{cvpr}              

\usepackage{graphicx}
\usepackage{amsmath}
\usepackage{amssymb}
\usepackage{booktabs}

\usepackage{color,colortbl}
\definecolor{ggray}{rgb}{0.92, 0.92, 0.99}
\newcolumntype{a}{>{\columncolor{ggray}}c}
\definecolor{awesome}{rgb}{1.0, 0.13, 0.32}
\usepackage{bbm}
\usepackage{algorithm,algorithmic}
\usepackage{enumitem}
\usepackage{flushend}
\usepackage[pagebackref,breaklinks,colorlinks]{hyperref}

\usepackage[capitalize]{cleveref}
\crefname{section}{Sec.}{Secs.}
\Crefname{section}{Section}{Sections}
\Crefname{table}{Table}{Tables}
\crefname{table}{Tab.}{Tabs.}

\usepackage{multirow}
\usepackage{setspace}
\usepackage[table,x11names,dvipsnames,table]{xcolor}
\def\cvprPaperID{***} 
\def\confName{CVPR}
\def\confYear{2022}

\begin{document}
	
	\title{DINE: Domain Adaptation from Single and Multiple Black-box Predictors}
	
	\author{Jian Liang
				 $^{1}$
				\qquad 
				Dapeng Hu $^2$
				\qquad
				Jiashi Feng $^{3}$
				\qquad
				Ran He $^{1,4,5}$\\
				$^1$ CRIPAC \& NLPR, CASIA
				\quad
				$^2$ NUS
				\quad
				$^3$ ByteDance
				\quad
				$^4$ UCAS
				\quad
				$^5$ CEBSIT, CAS \\
				{\tt\small liangjian92@gmail.com}\qquad
				{\tt\small dapeng.hu@u.nus.edu}\qquad
				{\tt\small jshfeng@gmail.com}\qquad
				{\tt\small rhe@nlpr.ia.ac.cn}
	}
	\maketitle
	
	\begin{abstract}
		To ease the burden of labeling, unsupervised domain adaptation (UDA) aims to transfer knowledge in previous and related labeled datasets (sources) to a new unlabeled dataset (target).
		Despite impressive progress, prior methods always need to access the raw source data and develop data-dependent alignment approaches to recognize the target samples in a transductive learning manner, which may raise privacy concerns from source individuals.
		Several recent studies resort to an alternative solution by exploiting the well-trained white-box model from the source domain, yet, it may still leak the raw data via generative adversarial learning.
		This paper studies a practical and interesting setting for UDA, where only black-box source models (\ie, only network predictions are available) are provided during adaptation in the target domain.
		To solve this problem, we propose a new two-step knowledge adaptation framework called DIstill and fine-tuNE (DINE).
		Taking into consideration the target data structure, DINE first distills the knowledge from the source predictor to a customized target model, then fine-tunes the distilled model to further fit the target domain.
		Besides, neural networks are not required to be identical across domains in DINE, even allowing effective adaptation on a low-resource device. 
		Empirical results on three UDA scenarios (\ie, single-source, multi-source, and partial-set) confirm that DINE achieves highly competitive performance compared to state-of-the-art data-dependent approaches. 
		Code is available at \url{https://github.com/tim-learn/DINE/}.
		
	\end{abstract}
	
	\section{Introduction}
	Deep neural networks achieve remarkable progress with massive labeled data, but it is expensive and not efficient to collect enough labeled data for each new task.
	To reduce the burden of labeling, considerable attention has been devoted to the transfer learning field \cite{pan2009survey,zhuang2020comprehensive}, especially for unsupervised domain adaptation (UDA) \cite{ben2007analysis,csurka2017domain}, where one or many related but different labeled datasets are collected as source domain(s) to help recognize unlabeled instances in the new dataset (called target domain).
	Recently, UDA methods have been widely applied in a variety of computer vision problems, \eg, image classification \cite{ganin2016domain,tzeng2017adversarial,long2018conditional}, semantic segmentation \cite{zhang2017curriculum,tsai2018learning,hu2021adversarial}, and object detection \cite{chen2018domain,khodabandeh2019robust,saito2019strong}. 
	
	\begin{figure}[t]
		\begin{center}
			\includegraphics[width=0.4\textwidth, trim=120 250 120 365,clip]{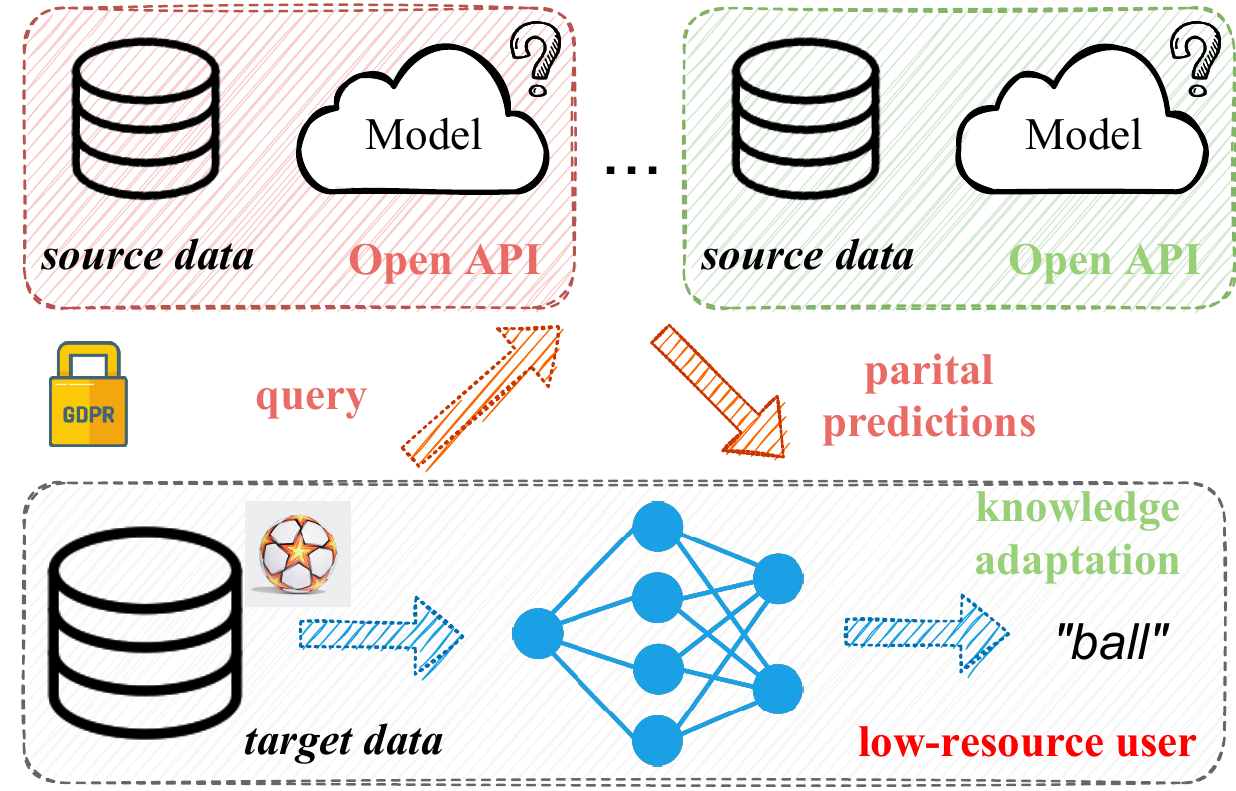}
			\caption{A challenging but interesting domain adaptation problem setting. One or many source agents only provide their black-box predictors (\eg, via the cloud API service) to the target user with certain unlabeled data, where neither \emph{the raw source data} nor \emph{the details about the source models} is accessible during adaptation.}
			\label{fig:framework}
		\end{center}
	\vspace{-15pt}
	\end{figure}
	
	Existing UDA methods always need to access the raw source data and resort to domain adversarial training \cite{tzeng2017adversarial,ganin2016domain} or maximum mean discrepancy minimization \cite{tzeng2014deep,long2015learning} to align source features with target features. 
	However, in many situations like personal medical records, the raw source data may be missing or must not be shared due to the privacy policy.
	To tackle this issue, several recent studies \cite{liang2020we,li2020model,kundu2020universal} attempt to utilize the trained model instead of the raw data from the source domain as supervision and obtain surprisingly good adaptation results.
	Even so, these methods always require source models to be elegantly trained and provided to the target domain with all the details, which raises two important concerns. 
	First, through generation techniques like generative adversarial learning \cite{goodfellow2014generative}, it is still possible to recover the raw source data, leaking the individual information. 
	Second, the target domain employs the same neural network as the source domain, which is not desirable for low-resource target users.
	Thus, this paper focuses on a realistic and challenging scenario for UDA, where the source model is trained without bells and whistles and provided to the target domain as a black-box predictor.
	
	For a better illustration, as shown in Fig.~\ref{fig:framework}, the target user exploits the API service offered by one or many source vendors to get the predictions for each instance and utilizes them for adaptation in the unlabeled target domain.
	To address such a challenging UDA problem, we propose a novel adaptation framework called DIstill and fine-tuNE (DINE).
	In a nutshell, DINE first distills the knowledge from predictions by source models and then fine-tunes the distilled model with the target data itself, forming a simple two-step approach.
	Note that, vanilla knowledge distillation \cite{hinton2015distilling} requires the existence of labeled data and learns the target model (student) by imitating the full outputs of the source model (teacher).
	Yet, besides the absence of labeled target data, acquiring the full teacher outputs is also impracticable in many situations, \eg, some predictors merely offer several highest soft-max probability and their associated labels.  
	To alleviate this issue, we devise an adaptive label smoothing technique on source predictions by keeping the largest soft-max value and forcing the rest with the same values. 
	
	On top of the point-wise supervision above, we introduce two kinds of structural regularizations into distillation for the first time: interpolation consistency training \cite{verma2019interpolation}---which encourages the prediction of interpolated samples to be consistent with the interpolated predictions; and mutual information maximization \cite{liang2020we,hu2017learning}---which helps increase the diversity among the target predictions.
	Thereafter, we aim to fit the target structure by adjusting parameters in the learned distilled model using the target data alone.
	For the sake of simplicity, we re-use mutual information maximization to fine-tune the distilled target model.
	As for multi-source UDA, we readily extend DINE by aggregating the outputs from multiple source predictors instead.
	We highlight the main contribution as follows: 
	\begin{itemize}
		\setlength{\itemsep}{0.1pt}
		\setlength{\parsep}{0.0pt}
		\setlength{\parskip}{0.0pt}
		\item We study a realistic and challenging UDA problem and propose a new adaptation framework (DINE) with only black-box predictors provided from source domains.
		\item We propose an adaptive label smoothing strategy and a structural distillation method by first introducing structural regularizations into unsupervised distillation.
		\item Empirical results on various benchmarks validate the superiority of the DINE framework over baselines. Provided with large source predictors like ViT \cite{dosovitskiy2020image}, DINE even yields state-of-the-art performance for single-source, multi-source, and partial-set UDA.
	\end{itemize}
	
	Compared to existing methods, DINE has several appealing aspects: 
	1) \textbf{safe}. It does not access the raw source data nor the source model, avoiding information leakage from source agents;
	2) \textbf{efficient}. It does not assume the same architecture across domains, so it can learn a lightweight target model from large source models. Moreover, it does not involve adversarial training \cite{li2020model} nor data synthesis \cite{kundu2020universal}, making the algorithm converge much faster.    
	
	\section{Related Work}
	\noindent\textbf{Domain Adaptation.}
	Domain adaptation uses labeled data in one or more source domains to solve new tasks in a target domain.
	This paper mainly focuses on a challenging problem---unsupervised domain adaptation (UDA), where no labeled data is available in the target domain.
	At early times, researchers address this problem via instance weighting \cite{huang2006correcting,sugiyama2007direct}, feature transformation \cite{pan2010domain,pan2010domain,liang2018aggregating}, and feature space \cite{gong2012geodesic,fernando2013unsupervised,sun2016return}.
	In the last decade, benefiting from representation learning, deep domain adaptation methods are prevailing and achieve remarkable progress. 
	To mitigate the gap between features across different domains, domain adversarial learning \cite{ganin2016domain,tzeng2017adversarial,long2018conditional,hoffman2018cycada} and discrepancy minimization \cite{tzeng2014deep,long2017deep,koniusz2017domain,kang2019contrastive} are widely used within deep UDA methods.
	Besides, another line of UDA methods \cite{saito2018maximum,chen2019domain,cui2020towards,jin2020minimum} focus on the network outputs and develop various regularization terms to pursue implicit domain alignment.
	In addition, researchers investigate other aspects of neural networks for UDA, \eg, domain-specific normalization-based methods \cite{maria2017autodial,chang2019domain} and feature regularization-based methods \cite{xu2019larger,chen2019transferability}.
	To fully verify the effectiveness, we study three UDA cases, \ie, single-source, multi-source \cite{peng2019moment,yang2020curriculum,li2021dynamic}, and partial-set (source label space subsumes target label space) \cite{cao2018partial,zhang2018importance,liang2020balanced}.
	
	\noindent\textbf{Model Transfer (Source data-free UDA).}
	Early parameter adaptation methods \cite{joachims1999transductive,duan2009domain} adapt the classifier trained in the source domain to the target domain with a small set of labeled examples, hence limiting their application in semi-supervised DA. 
	Besides empirical success, \cite{kuzborskij2013stability} pioneers the theoretical analysis of hypothesis transfer learning for linear regression.
	Inspired by this paradigm and increasingly important privacy concerns, \cite{chidlovskii2016domain,liang2019distant} develop several shallow adaptation methods without source data. 
	Several recent studies \cite{liang2020we,liang2021source,kundu2020universal,li2020model} introduce the source data-free setting for deep UDA, where the source domain merely offers a trained model.
	Specifically, \cite{liang2020we} freezes the classifier layer and fine-tunes the feature module via information maximization and pseudo-labeling in the target domain, which is further extended to multi-source UDA \cite{ahmed2021unsupervised,feng2021kd3a}. 
	\cite{li2020model} leverages a conditional generative adversarial net and incorporates generated images into the adaptation process.
	However, exposing details of the trained source model is fairly risky due to some committed white-box attacks.
	Faced with a black-box source model, \cite{liang2021source} divides the target dataset into two splits and employs semi-supervised learning to enhance the performance of the uncertain split.
	\cite{lipton2018detecting} focuses on black-box label shift, but it requires a hold-out source set to estimate class confusion matrix, which is sometimes hard to satisfy.
	Moreover, \cite{zhang2021unsupervised} proposes an iterative noisy label learning approach using full soft labels.
	DINE focuses on the black-box covariate shift problem and works well even only hard labels from source predictors provided.
	
	\begin{figure*}[t]
		\begin{center}
			\includegraphics[width=0.85\textwidth, trim=0 0 0 0,clip]{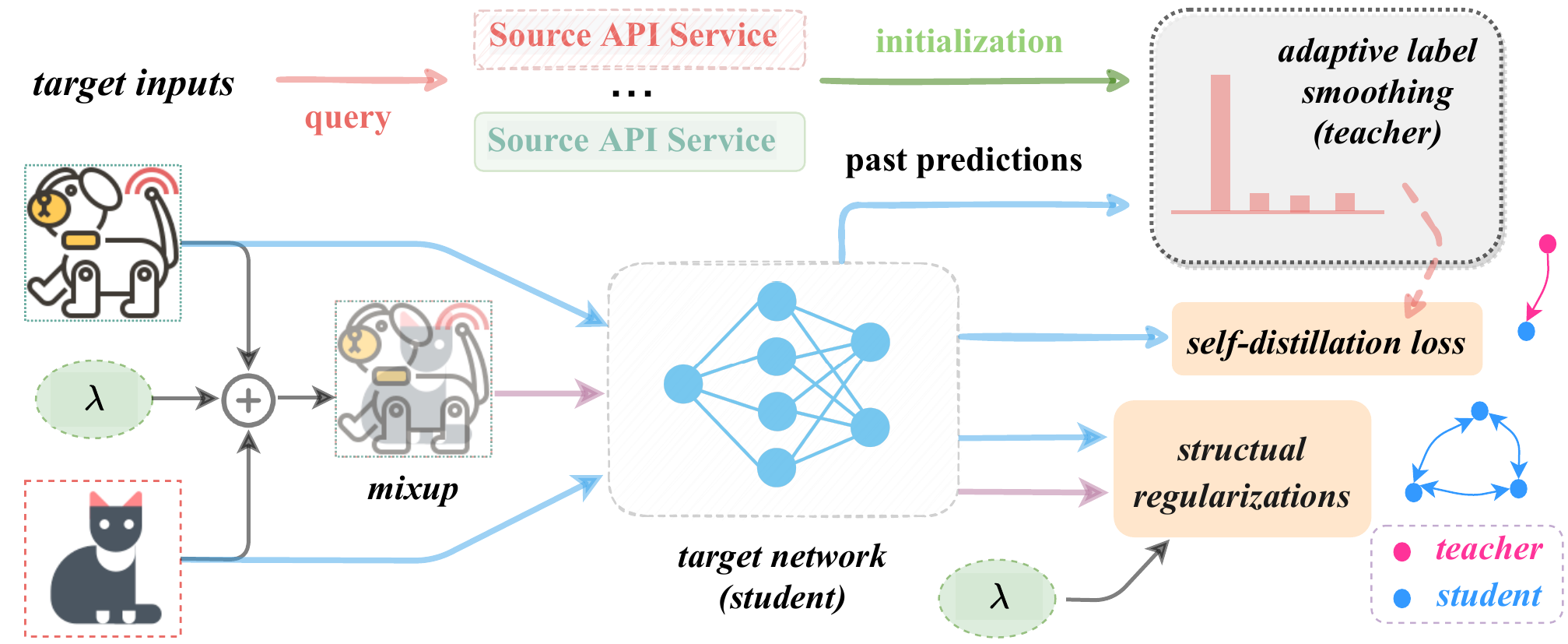}
			\caption{An overview of the proposed DINE framework. Black-box source predictors (\ie, API service) are merely required to initialize the memory bank that stores the predictions of each target instance. In the self-distillation loss, the memory bank could be considered as a teacher that maintains an exponential moving average (EMA) prediction. Structural regularizations, describing batch-wise and pair-wise data structure, are further employed during adaptation. During the fine-tuning step, only the mutual information objective is activated.}
			\label{fig:distune}
		\end{center}
	\vspace{-10pt}
	\end{figure*}
	
	\noindent\textbf{Knowledge Distillation.}
	Knowledge distillation aims to transfer knowledge from one model (\textit{a.k.a.}, teacher) to another model (\textit{a.k.a.}, student), usually from a larger one to a smaller one. 
	\cite{hinton2015distilling} shows that, augmenting the training of the student with a distillation loss, matching the predictions between teacher and student, is beneficial.
	In fact, knowledge distillation could be considered as a learned label smoothing regularization \cite{szegedy2016rethinking}, which has a similar function with the latter \cite{yuan2020revisiting}.
	Recently, \cite{kim2021self} proposes self-knowledge distillation and shows that, even within the same neural network, the past predictions could be the teacher itself.
	Besides supervised training, such self-distillation could be effectively applied with unlabeled data for semi-supervised learning.
	For instance, \cite{laine2016temporal} proposes to ensemble the predictions during training, using the outputs of a single network on different training epochs, as a teacher for the current epoch.
	Instead of maintaining an exponential moving average (EMA) prediction in \cite{laine2016temporal}, \cite{tarvainen2017mean} utilizes an average of consecutive student models (past model weights) as a stronger teacher, but is not applicable for black-box UDA here.
	DINE proposes an adaptive label smoothing technique on source predictions and for the first time introduces structural regularizations \cite{verma2019interpolation,gomes2010discriminative} into unsupervised distillation.
	
	\section{Methodology}
	We mainly focus on the $K$-way cross-domain image classification task and aim to address an interesting, realistic but challenging UDA setting, where only single or multiple black-box source predictors are provided to the unlabeled target domain.
	For the single-source UDA scenario, the source domain $\{x_s^i, y_s^i\}_{i=1}^{n_s}$ consists of $n_s$ labeled instances, where $x_s^i \in \mathcal{X}_s, y_s^i \in \mathcal{Y}_s$, and the target domain $\{x_t^i, y_t^i\}_{i=1}^{n_t}$ consists of $n_t$ unlabeled instances, where $x_t^i \in \mathcal{X}_t, y_t^i \in \mathcal{Y}_t$, and the goal is to infer the values of $\{y_t^i\}_{i=1}^{n_t}$.
	The label spaces are always assumed the same across domains, \ie, $\mathcal{Y}_s = \mathcal{Y}_t$, for single-source and multi-source UDA.
	By contrast, partial-set UDA \cite{cao2018partial} assumes that some source classes does not exist in the target domain, \ie, $\mathcal{Y}_s \supset \mathcal{Y}_t$.
	Concerning the black-box adaptation setting, only these trained source models are provided, without requiring access to the source data. 
	It also differs from prior model adaptation methods \cite{liang2020we,li2020model} in requiring no details about the source model, \eg, backbone type and network parameters.
	In particular, \emph{only the hard or soft network predictions of the target instances $\mathcal{X}_t$ from the source model $f_s: \mathcal{X}_s\to \mathcal{Y}_s$} are utilized for single-source adaptation in the target domain.
	
	\subsection{Architecture and Source Model Generation}
	We elaborate on how to obtain the trained model from the source domain as follows. Unlike \cite{liang2020we,liang2021source} that elegantly design the source model with a bottleneck layer and the weight normalization \cite{salimans2016weight} technique, we just insert a single linear fully-connected (FC) layer after the backbone feature network, and use label smoothing (LS) \cite{szegedy2016rethinking} to train $f_s$,
	\begin{equation}
		\mathcal{L}_{s}(f_s;\mathcal{X}_s,\mathcal{Y}_s) = -
		\mathbb{E}_{(x_s,y_s)\in \mathcal{X}_s \times \mathcal{Y}_s}\ (q^s)^T \log f_s(x_s),
		\label{eq:ls}
	\end{equation} 
	where $q^s=(1-\alpha)\mathbf{1}_{y_s} + \alpha/K$ is the smoothed label vector and $\alpha$ is the smoothing parameter empirically set to 0.1, and $\mathbf{1}_{j}$ denotes a $K$-dimensional one-hot encoding with only the $j$-th value being 1.
	
	As for the self-defined target network $f_t: \mathcal{X}_t\to \mathcal{Y}_t$, we follow \cite{liang2020we,liang2021source} and adopt the bottleneck layer consisting of a batch-normalization layer and an FC layer, and the classifier consisting of a weight-normalization layer and an FC layer.
	
	\subsection{Adaptive Self-Knowledge Distillation}
	In order to extract knowledge from a black-box model, there exists a natural solution called knowledge distillation (KD) \cite{hinton2015distilling} by forcing the target model (student) to learn similar predictions to the source model (teacher). However, existing KD methods are applied to a supervised training task, and the consistency loss in the following works well by acting as a regularization term,
	\begin{equation}
		\mathcal{L}_{kd}(f_t;\mathcal{X}_t, f_s) =
		\mathbb{E}_{x_t\in \mathcal{X}_t} \mathcal{D}_{kl}\left(f_s(x_t) \ ||\ f_t(x_t)\right),
		\label{eq:kd}
	\end{equation} 
	where $D_{kl}$ denotes the Kullback-Leibler (KL) divergence loss. 
	However, the network outputs from the source model $f_s$ for target instances are \emph{not accurate and sometimes even incomplete}. 
	For the studied black-box adaptation problem, highly relying on the teacher $f_s(x_t)$ via a consistency loss above sounds not desirable anymore.
	Thus, we propose to revise the teacher output $p$ via top-$r$ largest values as,
	\begin{equation}
		\text{AdaLS}(p,r)_i =\left\{
		\begin{aligned}
			p_i, \qquad \qquad \qquad \qquad \qquad   \; &  i \in \mathcal{T}^r_p, \\
			(1 - \sum\nolimits_{j \in \mathcal{T}^r_p} p_j)/ (K-r), \; & \text{otherwise}.
		\end{aligned}
		\right.
		\label{eq:als}
	\end{equation} 	
	Here $\mathcal{T}^r_p$ denotes the index set of top-$r$ classes in $p$, and we term this transformation in Eq.~(\ref{eq:als}) as adaptive label smoothing (\textbf{adaptive LS}), since these instance-specific top-$r$ values are kept which are not the same for different samples. 
	As a byproduct, using the smoothed output ($r=1$) means that we merely need the predicted class along with its maximum probability, which sounds more flexible when using an API service provided by other companies.	
	The refined output $\text{AdaLS}(p,r)$ is believed to work better than the original output $p$ for several reasons below, 1) it partially neglects some redundant and noisy information by only focusing on the pseudo label (class associated with the largest value) and forcing a uniform distribution on other classes like label smoothing \cite{szegedy2016rethinking}; 2) it does not solely rely on the noisy pseudo label but utilizes the largest value as confidence, similar to self-weighted pseudo labeling \cite{iscen2019label}.
	Generally, we first obtain the smoothed predictions from single or multiple source predictors as the initialized teacher as,
	\begin{equation}
		P^T(x_t) \gets \frac{1}{M} \sum\nolimits_{m=1}^{M} \text{AdaLS}(f_s^{(m)}(x_t)),
		\label{eq:init}
	\end{equation} 	
	where $M$ denotes the number of predictors, $f_s^{(m)}(x_t)$ denotes the predictions of $x_t$ through $m$-th source predictor. 
	
	To further alleviate the noise in the teacher prediction, we follow \cite{laine2016temporal,kim2021self} and adopt a self-distillation strategy, shown in Fig.~\ref{fig:distune}, maintaining an EMA prediction by 
	\begin{equation}
		P^T(x_t) \gets \gamma P^T(x_t) + (1-\gamma) f_t(x_t), \ \forall x_t \in \mathcal{X}_t,
		\label{eq:ema}
	\end{equation}
	where $\gamma$ is a momentum hyper-parameter.
	Following \cite{laine2016temporal}, we update teacher predictions after every training epoch.
	When $\gamma=1$, there exists no temporal ensembling, \ie, the source predictions act as a teacher throughout distillation.	
	
	\subsection{Distillation with Structural Regularizations}
	As stated above, the teacher output from the source model is highly possible to be inaccurate and noisy due to the domain shift.
	Even we devise a promising solution in Eq.~(\ref{eq:init}), only the point-wise information is considered during the distillation process, it ignores the data structure in the target domain, thus not enough for effective noisy knowledge distillation.
	As such, we incorporate the structural information in the target domain to regularize the distillation. 
	First, we consider the pairwise structural information via MixUp \cite{zhang2017mixup}, and employ the interpolation consistency training \cite{verma2019interpolation} technique as below,
	\begin{equation}
		\begin{aligned}
			\mathcal{L}_{mix} & (f_t;\mathcal{X}_t) =
			\mathbb{E}_{x^t_i, x^t_j\in \mathcal{X}_t} \mathbb{E}_{\lambda \in \text{Beta}(\alpha,\alpha)} \\
			& l_{ce}\left(
			\text{Mix}_{\lambda}\left(f'_t(x^t_i), f'_t(x^t_j)\right), f_t\left(\text{Mix}_{\lambda}(x^t_i, x^t_j)\right)
			\right), \\
		\end{aligned}
		\label{eq:mix}
	\end{equation}
	where $l_{ce}$ denotes the cross-entropy loss,  and $\text{Mix}_{\lambda}(a,b)=\lambda\cdot a + (1-\lambda)\cdot b$ denotes the MixUp operation, and $\lambda$ is sampled from a Beta distribution, and $\alpha$ is the hyper-parameter empirically set to 0.3 \cite{zhang2017mixup}.
	$f'_t$ just offers the values of $f_t$ but needs no gradient optimization. Here we do not adopt the EMA update strategy in \cite{verma2019interpolation} for $f'_t$.
	Eq.~(\ref{eq:mix}) can be treated to augment the target domain with more interpolated samples, which is beneficial for better generalization ability. 
	
	In addition, we also consider the global structural information during distillation in the target domain.
	In fact, during distillation, the classes with a large number of instances are relatively easy to learn, which may wrongly recognize some confusing target instances as such classes in turn. 
	To circumvent this problem, we attempt to encourage diversity among the predictions of all the target instances.
	Specifically, we try to maximize the widely-used mutual information objective \cite{gomes2010discriminative,hu2017learning,liang2020we} in the following,
	\begin{equation}
		\begin{aligned}
			\mathcal{L}_{im}(f_t;\mathcal{X}_t) & = 
			H(\mathcal{Y}_t) - H(\mathcal{Y}_t | \mathcal{X}_t) \\
			& = h\left(\mathbb{E}_{x_t\in\mathcal{X}_t} f_t(x_t)\right) - \mathbb{E}_{x_t\in\mathcal{X}_t}\ h\left(f_t(x_t)\right), 
		\end{aligned}
		\label{eq:im}
	\end{equation}
	where $h(p)=-\sum_i p_i \log p_i$ represents the conditional entropy function.
	Note that, increasing the marginal entropy $H(\mathcal{Y}_t)$ encourages the label distribution to be uniform while decreasing the conditional entropy $H(\mathcal{Y}_t|\mathcal{X}_t)$ encourages unambiguous network predictions.
	
	Integrating these objectives introduced in Eqs.~(\ref{eq:kd}, \ref{eq:mix}, \ref{eq:im}) together,  we obtain the final loss function as follows,
	\begin{equation}
		\mathcal{L}_{t} = \mathcal{D}_{kl}\left(P^T(x_t) \ ||\ f_t(x_t)\right) + \beta \mathcal{L}_{mix} - \mathcal{L}_{im},\\
		\label{eq:overall}
	\end{equation}
	where $\beta$ is a hyper-parameter empirically set to 1, controlling the importance of $\mathcal{L}_{mix}$ during structural distillation.
	
	Different from the most closely related work \cite{zhang2021unsupervised} that iteratively refines the pseudo labels and optimizes the target network, DINE directly learns good network predictions for the target data as a unified approach, which is more desirable to capture the data structure of the target domain. 
	
	\subsection{Fine-tuning the Distilled Model}
	Through the proposed structural knowledge distillation method from black-box source predictors $\{f_s^{(m)}\}_{m=1}^{M}$, it is expected to learn a well-performing white-box target model. 
	However, the distilled model seems sub-optimal since it is mainly optimized via the point-wise knowledge distillation term in Eq.~(\ref{eq:kd}), which highly depends on the source predictions.
	Inspired by DIRT-T \cite{shu2018dirt}, we hypothesize that a better network is achievable by introducing a secondary training phase that solely minimizes the target-side cluster assumption violation. 
	Rather than employ the parameter-sensitive virtual adversarial training \cite{shu2018dirt}, we again employ the mutual information maximization in Eq.~(\ref{eq:im}) to refine the distilled target model.
	So far, we have shown all the details of two steps within the proposed framework (DINE). 
	A full description of DINE is further provided in Algorithm~\ref{alg:DINE}.
	
	\section{Experiments}
	\subsection{Setup}
	\noindent \textbf{a) Datasets.} 
	\textbf{Office} \cite{saenko2010adapting} is a popular benchmark on cross-domain object recognition, consisting of three different domains in 31 categories. 
	\textbf{Office-Home} \cite{venkateswara2017deep} is a challenging medium-sized benchmark on object recognition, consisting of four different domains in 65 categories.
	\textbf{VisDA-C} \cite{peng2017visda} is a large-scale benchmark developed for 12-class synthetic-to-real object recognition. The source domain contains 152 thousand synthetic images generated by rendering 3D models while the target domain has 55 thousand real object images from Microsoft COCO.
	Gong et al. \cite{gong2012geodesic} further extract 10 shared categories between Office and Caltech-256 to form a new benchmark named \textbf{Office-Caltech}.
	\textbf{Image-CLEF} is a benchmark for ImageCLEF 2014 domain adaptation challenge \footnote{\url{http://imageclef.org/2014/adaptation}}, organized by selecting the
	12 common categories shared by three public datasets.
	
	\setlength{\tabcolsep}{1.0pt}
	\begin{table}[]
		\centering
		\caption{Accuracies (\%) on \textbf{Office} \cite{saenko2010adapting} for single-source closed-set UDA. (Best value of source-prediction-based (Pred.) methods in \textbf{\color{awesome}{bold}}),  `Mod.' denotes source-model-based, `Data' denotes source-data-dependent. \textbf{* denotes ViT-based}.}
		\resizebox{0.475\textwidth}{!}{$
			\begin{tabular}{lccccccca}
				\toprule
				Method & Type & A$\to$D & A$\to$W & D$\to$A & D$\to$W & W$\to$A & W$\to$D & Avg. \\
				\midrule
				No Adapt. & Pred. & 79.9 & 76.6 & 56.4 & 92.8 & 60.9 & 98.5 & 77.5 \\
				NLL-OT \cite{asano2019self} & Pred. & 88.8 & 85.5 & 64.6 & 95.1 & 66.7 & 98.7 & 83.2 \\
				NLL-KL \cite{zhang2021unsupervised} & Pred. & 89.4 & 86.8 & 65.1 & 94.8 & 67.1 & 98.7 & 83.6 \\
				HD-SHOT \cite{liang2020we} & Pred. & 86.5 & 83.1 & 66.1 & 95.1 & 68.9 & 98.1 & 83.0 \\
				SD-SHOT \cite{liang2020we} & Pred. & 89.2 & 83.7 & 67.9 & 95.3 & 71.1 & 97.1 & 84.1 \\
				DINE & Pred. & 91.6 & 86.8 & 72.2 & 96.2 & 73.3 & 98.6 & 86.4 \\
				DINE (full) & Pred. & 91.7 & 87.5 & 72.9 & 96.3 & 73.7 & 98.5 & 86.7 \\
				\midrule
				\multicolumn{9}{c}{\textbf{ResNet-50$\upuparrows$}, \textbf{ViT$\downdownarrows$} (source backbone) $\to$ \textbf{ResNet-50} (target backbone)} \\
				\midrule
				No Adapt. & Pred. & 88.2 & 89.2 & 74.5 & 97.2 & 77.2 & 99.3 & 87.6 \\
				NLL-OT \cite{asano2019self} & Pred. & 91.3 & 91.4 & 76.4 & 97.2 & 78.2 & 99.4 & 89.0 \\
				NLL-KL \cite{zhang2021unsupervised} & Pred. & 91.7 & 91.8 & 76.3 & 97.2 & 78.4 & 99.0 & 89.1 \\
				HD-SHOT \cite{liang2020we} & Pred. & 88.8 & 90.9 & 75.3 & 97.7 & 77.7 & 99.5 & 88.3 \\
				SD-SHOT \cite{liang2020we} & Pred. & 91.6 & 92.8 & 77.8 & 98.7 & 78.5 & \textbf{\color{awesome}99.7} & 89.8 \\
				DINE & Pred. & 94.2 & 94.6 & 80.7 & \textbf{\color{awesome}98.8} & 81.5 & 99.5 & 91.6 \\
				DINE (full) & Pred. & \textbf{\color{awesome}95.5} & \textbf{\color{awesome}94.8} & \textbf{\color{awesome}81.2} & 98.5 & \textbf{\color{awesome}82.0} & \textbf{\color{awesome}99.7} & \textbf{\color{awesome}91.9} \\
				\midrule
				\midrule
				SHOT \cite{liang2021source} & Mod. & 93.9 & 90.1 & 75.3 & 98.7 & 75.0 & 99.9 & 88.8 \\
				SHOT++ \cite{liang2021source} & Mod. & 94.5 & 90.9 & 76.3 & 98.6 & 75.8 & 99.9 & 89.3 \\
				A$^2$Net \cite{xia2021adaptive} & Mod. & 94.5 & 94.0 & 76.7 & 99.2 & 76.1 & 100. & 90.1 \\
				TransDA* \cite{yang2021transformer} & Mod. & 97.2 & 95.0 & 73.7 & 99.3 & 79.3 & 99.6 & 90.7 \\
				SCDA$_{\text{MDD}}$ \cite{li2021semantic} & Data & 95.4 & 95.3 & 77.2 & 99.0 & 75.9 & 100. & 90.5 \\
				SRDC \cite{tang2020unsupervised} & Data & 95.8 & 95.7 & 76.7 & 99.2 & 77.1 & 100. & 90.8 \\
				RADA$_{\text{CDAN}}$ \cite{jin2021re} & Data & 96.1 & 96.2 & 77.5 & 99.3 & 77.4 & 100. & 91.1 \\
				\bottomrule
			\end{tabular}
			$}
		\label{tab:office}
	\end{table}
	
	\setlength{\tabcolsep}{3.0pt}
	\begin{table*}[]
		\centering
		\caption{Accuracies (\%) on \textbf{Office-Home} \cite{venkateswara2017deep} for single-source closed-set UDA.}
		\vspace{-5pt}
		\resizebox{0.88\textwidth}{!}{$
			\begin{tabular}{lccccccccccccca}
				\toprule
				Method & Type & Ar$\to$Cl & Ar$\to$Pr & Ar$\to$Re & Cl$\to$Ar & Cl$\to$Pr & Cl$\to$Re & Pr$\to$Ar & Pr$\to$Cl & Pr$\to$Re & Re$\to$Ar & Re$\to$Cl & Re$\to$Pr & Avg. \\
				\midrule
				No Adapt. & Pred. & 44.1 & 66.9 & 74.2 & 54.5 & 63.3 & 66.1 & 52.8 & 41.2 & 73.2 & 66.1 & 46.7 & 77.5 & 60.6 \\
				NLL-OT \cite{asano2019self} & Pred. & 49.1 & 71.7 & 77.3 & 60.2 & 68.7 & 73.1 & 57.0 & 46.5 & 76.8 & 67.1 & 52.3 & 79.5 & 64.9 \\
				NLL-KL \cite{zhang2021unsupervised} & Pred. & 49.0 & 71.5 & 77.1 & 59.0 & 68.7 & 72.9 & 56.4 & 46.9 & 76.6 & 66.2 & 52.3 & 79.1 & 64.6 \\
				HD-SHOT \cite{liang2020we} & Pred. & 48.6 & 72.8 & 77.0 & 60.7 & 70.0 & 73.2 & 56.6 & 47.0 & 76.7 & 67.5 & 52.6 & 80.2 & 65.3 \\
				SD-SHOT \cite{liang2020we} & Pred. & 50.1 & 75.0 & 78.8 & 63.2 & 72.9 & 76.4 & 60.0 & 48.0 & 79.4 & 69.2 & 54.2 & 81.6 & 67.4 \\
				DINE & Pred. & 52.2 & 78.4 & 81.3 & 65.3 & 76.6 & 78.7 & 62.7 & 49.6 & 82.2 & 69.8 & 55.8 & 84.2 & 69.7\\
				DINE (full) & Pred. & 54.2 & 77.9 & 81.6 & 65.9 & 77.7 & 79.9 & 64.1 & 50.5 & 82.1 & 71.1 & 58.0 & 84.3 & 70.6 \\
				\midrule
				\multicolumn{15}{c}{\textbf{ResNet-50$\uparrow$}, \textbf{ViT$\downarrow$} (source backbone) $\to$ \textbf{ResNet-50} (target backbone)} \\
				\midrule
				No Adapt. & Pred. & 54.5 & 83.2 & 87.2 & 78.0 & 83.8 & 86.1 & 74.5 & 49.7 & 87.4 & 78.6 & 52.6 & 86.2 & 75.1 \\
				NLL-OT \cite{asano2019self} & Pred. & 58.8 & 84.4 & 87.6 & 78.2 & 84.7 & 86.7 & 76.0 & 54.0 & 88.0 & 79.7 & 57.2 & 87.2 & 76.9 \\
				NLL-KL \cite{zhang2021unsupervised} & Pred. & 59.5 & 84.3 & 87.6 & 77.4 & 84.8 & 86.8 & 75.1 & 54.9 & 88.0 & 79.0 & 57.9 & 87.2 & 76.9 \\
				HD-SHOT \cite{liang2020we} & Pred. & 57.2 & 84.2 & 87.3 & 78.4 & 84.9 & 86.4 & 74.8 & 56.0 & 87.6 & 78.9 & 57.5 & 87.0 & 76.7 \\
				SD-SHOT \cite{liang2020we} & Pred. & 59.4 & 85.2 & 87.8 & 79.6 & 86.6 & 87.1 & 76.4 & 58.3 & 87.8 & 80.0 & 59.5 & 87.9 & 78.0 \\
				DINE & Pred. & \textbf{\color{awesome}64.9} & 87.4 & 88.8 & 80.5 & \textbf{\color{awesome}89.6} & 87.8 & 79.0 & \textbf{\color{awesome}62.9} & 89.1 & 81.5 & 64.6 & \textbf{\color{awesome}90.0} & 80.5 \\
				DINE (full) & Pred. & 64.4 & \textbf{\color{awesome}87.9} & \textbf{\color{awesome}89.0} & \textbf{\color{awesome}80.9} & \textbf{\color{awesome}89.6} & \textbf{\color{awesome}88.7} & \textbf{\color{awesome}79.6} & 62.5 & \textbf{\color{awesome}89.4} & \textbf{\color{awesome}81.7} & \textbf{\color{awesome}65.2} & 89.7 & \textbf{\color{awesome}80.7} \\
				\midrule
				\midrule
				SHOT \cite{liang2021source} & Mod. & 57.7 & 79.1 & 81.5 & 67.6 & 77.9 & 77.8 & 68.1 & 55.8 & 82.0 & 72.8 & 59.7 & 84.4 & 72.0\\
				A$^2$Net \cite{xia2021adaptive} & Mod. & 58.4 & 79.0 & 82.4 & 67.5 & 79.3 & 78.9 & 68.0 & 56.2 & 82.9 & 74.1 & 60.5 & 85.0 & 72.8 \\
				SHOT++ \cite{liang2021source} & Mod. & 58.1 & 79.5 & 82.4 & 68.6 & 79.9 & 79.3 & 68.6 & 57.2 & 83.0 & 74.3 & 60.4 & 85.1 & 73.0 \\
				TransDA* \cite{yang2021transformer} & Mod. & 67.5 & 83.3 & 85.9 & 74.0 & 83.8 & 84.4 & 77.0 & 68.0 & 87.0 & 80.5 & 69.9 & 90.0 & 79.3 \\
				RADA$_{\text{CDAN}}$ \cite{jin2021re} & Data & 56.5 & 76.5 & 79.5 & 68.8 & 76.9 & 78.1 & 66.7 & 54.1 & 81.0 & 75.1 & 58.2 & 85.1 & 71.4 \\
				ATDOC-NA \cite{liang2021domain} & Data & 58.3 & 78.8 & 82.3 & 69.4 & 78.2 & 78.2 & 67.1 & 56.0 & 82.7 & 72.0 & 58.2 & 85.5 & 72.2 \\
				SCDA$_{\text{DCAN}}$ \cite{li2021semantic} & Data & 60.7 & 76.4 & 82.8 & 69.8 & 77.5 & 78.4 & 68.9 & 59.0 & 82.7 & 74.9 & 61.8 & 84.5 & 73.1 \\
				\bottomrule
			\end{tabular}
			$}
		\label{tab:home}
		\vspace{-5pt}
	\end{table*}
	
	\noindent \textbf{b) Implementation details.} 
	Generally, we randomly run our methods three times with different random seeds \{2019, 2020, 2021\} via \textbf{PyTorch} and report the average accuracies. 
	Regarding the source model $f_s$, we train it using all the samples in the source domain.
	In this paper, we mainly consider three different backbones, ResNet-50, ResNet-101 \cite{he2016deep} and ViT-B\_16 \cite{dosovitskiy2020image} (ViT for simplicity).
	Following \cite{liang2020we}, mini-batch SGD is employed to learn the layers initialized from the ImageNet pre-trained model or last stage with the learning rate (1e-3), and new layers from scratch with the learning rate (1e-2). 
	Besides, we use the suggested training settings in \cite{long2018conditional,liang2020we}, including learning rate scheduler, momentum (0.9), weight decay (1e-3), bottleneck size (256), and batch size (64).
	Concerning the parameters in DINE, $r=1$ and $T_m=30$ are adopted for all datasets and tasks except $T_m=10$ for \textbf{VisDA-C}. 
	Moreover, two hyper-parameters $\beta=1.0, \gamma=0.6$ are fixed throughout this paper.
	
	\setlength{\tabcolsep}{5.0pt} 
	\begin{table*}[]
		\centering
		\caption{Accuracies (\%) on \textbf{VisDA-C} \cite{peng2017visda} for single-source closed-set UDA.}
		\vspace{-5pt}
		\resizebox{0.88\textwidth}{!}{$
			\begin{tabular}{lccccccccccccca}
				\toprule
				Method & Type & plane & bcycl & bus & car & horse & knife & mcycle & person & plant & sktbrd & train & truck & Per-class \\
				\midrule
				No Adapt. & Pred. & 64.3 & 24.6 & 47.9 & 75.3 & 69.6 & 8.5 & 79.0 & 31.6 & 64.4 & 31.0 & 81.4 & 9.2 & 48.9 \\
				NLL-OT \cite{asano2019self} & Pred. & 82.6 & 84.1 & 76.2 & 44.8 & 90.8 & 39.1 & 76.7 & 72.0 & 82.6 & 81.2 & 82.7 & 50.6 & 72.0 \\
				NLL-KL \cite{zhang2021unsupervised} & Pred. & 82.7 & 83.4 & 76.7 & 44.9 & 90.9 & 38.5 & 78.4 & 71.6 & 82.4 & 80.3 & 82.9 & 50.4 & 71.9 \\
				HD-SHOT \cite{liang2020we} & Pred. & 75.8 & 85.8 & 78.0 & 43.1 & 92.0 & 41.0 & 79.9 & 78.1 & 84.2 & 86.4 & 81.0 & 65.5 & 74.2 \\
				SD-SHOT \cite{liang2020we} & Pred. & 79.1 & 85.8 & 77.2 & 43.4 & 91.6 & 41.0 & 80.0 & 78.3 & 84.7 & 86.8 & 81.1 & 65.1 & 74.5 \\
				DINE & Pred. & 81.4 & 86.7 & 77.9 & 55.1 & 92.2 & 34.6 & 80.8 & 79.9 & 87.3 & 87.9 & 84.3 & 58.7 & 75.6 \\
				DINE (full) & Pred. & 95.3 & 85.9 & 80.1 & 53.4 & 93.0 & 37.7 & 80.7 & 79.2 & 86.3 & 89.9 & 85.7 & 60.4 & 77.3 \\
				\midrule
				\multicolumn{15}{c}{source backbone: \textbf{ResNet-101 ($\upuparrows$)}, \textbf{ViT ($\downdownarrows$)} $\to$ \textbf{ResNet-101} (target backbone)} \\
				\midrule
				No Adapt. & Pred. & 97.0 & 56.2 & 81.0 & \textbf{\color{awesome}74.4} & 91.8 & 52.0 & 92.5 & 10.1 & 73.4 & 92.7 & \textbf{\color{awesome}97.0} & 17.5 & 69.6 \\
				NLL-OT \cite{asano2019self} & Pred. & \textbf{\color{awesome}97.8} & 90.8 & 81.9 & 49.7 & 95.7 & 93.5 & 85.2 & 45.4 & 88.9 & \textbf{\color{awesome}96.6} & 91.2 & 54.4 & 80.9 \\
				NLL-KL \cite{zhang2021unsupervised} & Pred. & 97.6 & 91.1 & 82.1 & 49.2 & \textbf{\color{awesome}95.8} & 93.5 & \textbf{\color{awesome}86.2} & 44.6 & 89.0 & 96.4 & 91.4 & 54.8 & 81.0 \\
				HD-SHOT \cite{liang2020we} & Pred. & 96.7 & 91.7 & 81.8 & 48.4 & 95.1 & \textbf{\color{awesome}98.5} & 83.1 & 60.1 & \textbf{\color{awesome}92.2} & 87.7 & 88.4 & \textbf{\color{awesome}65.3} & 82.4 \\
				SD-SHOT \cite{liang2020we} & Pred. & 96.3 & 91.1 & 80.3 & 46.4 & 93.9 & 98.2 & 81.5 & 58.6 & 90.9 & 85.5 & 88.0 & 63.8 & 81.2 \\
				DINE & Pred. & 96.6 & \textbf{\color{awesome}91.9} & \textbf{\color{awesome}83.1} & 58.2 & 95.3 & 97.8 & 85.0 & \textbf{\color{awesome}73.6} & 91.9 & 94.9 & 92.2 & 60.7 & \textbf{\color{awesome}85.1} \\
				DINE (full) & Pred. & 96.6 & \textbf{\color{awesome}91.9} & 82.9 & 57.9 & 95.4 & 97.8 & 84.5 & 73.1 & 91.7 & 95.1 & 92.0 & 60.9 & 85.0 \\
				\midrule
				\midrule
				A$^2$Net \cite{xia2021adaptive} & Mod. & 94.0 & 87.8 & 85.6 & 66.8 & 93.7 & 95.1 & 85.8 & 81.2 & 91.6 & 88.2 & 86.5 & 56.0 & 84.3 \\
				SHOT \cite{liang2021source} & Mod. & 95.8 & 88.2 & 87.2 & 73.7 & 95.2 & 96.4 & 87.9 & 84.5 & 92.5 & 89.3 & 85.7 & 49.1 & 85.5 \\
				SHOT++ \cite{liang2021source} & Mod. & 95.8 & 88.3 & 90.5 & 84.7 & 97.9 & 98.0 & 92.9 & 85.3 & 97.5 & 92.9 & 93.9 & 32.3 & 87.5 \\
				TransDA* \cite{yang2021transformer} & Mod. & 97.2 & 91.1 & 81.0 & 57.5 & 95.3 & 93.3 & 82.7 & 67.2 & 92.0 & 91.8 & 92.5 & 54.7 & 83.0 \\
				ATDOC-NA \cite{liang2021domain} & Data & 93.7 & 83.0 & 76.9 & 58.7 & 89.7 & 95.1 & 84.4 & 71.4 & 89.4 & 80.0 & 86.7 & 55.1 & 80.3 \\
				STAR \cite{lu2020stochastic} & Data & 95.0 & 84.0 & 84.6 & 73.0 & 91.6 & 91.8 & 85.9 & 78.4 & 94.4 & 84.7 & 87.0 & 42.2 & 82.7 \\
				CAN \cite{kang2019contrastive} & Data & 97.0 & 87.2 & 82.5 & 74.3 & 97.8 & 96.2 & 90.8 & 80.7 & 96.6 & 96.3 & 87.5 & 59.9 & 87.2 \\
				\bottomrule
			\end{tabular}
			$}
		\label{tab:visda}
		\vspace{-10pt}
	\end{table*}
	
	\noindent \textbf{c) Baselines.}
	Since the black-box UDA setting is fairly new in this field, we come up with several baseline methods below.
	\textbf{No Adapt.} is also known as `source only' in this field that infers the class label from the predictions. 
	\textbf{NLL-KL} mainly follows the idea of noisy label learning (NLL) \cite{zhang2021unsupervised} and adopts the diversity-promoting KL divergence to refine the noisy pseudo labels, then train a network with the refined pseudo labels iteratively.
	\textbf{NLL-OT} differs from NLL-KL only in that the optimal transport (OT) technique \cite{asano2019self} is employed instead of the KL divergence in the refining step.
	\textbf{HD-SHOT} first learns a white-box model by self-training in the target domain and then exploits SHOT \cite{liang2020we} for further adaptation.
	It treats the class predicted by $f_s$ as the true label for each target instance and employs a cross-entropy loss to train the model. 
	\textbf{SD-SHOT} differs from HD-SHOT only in that a weighted cross-entropy loss is utilized where the predictive confidence is treated as the instance weight. 
	
	We provide DINE (w/o FT) and DINE (full) besides DINE. 
	When dropping the second fine-tuning step, DINE becomes DINE (w/o FT). 
	DINE becomes DINE (full) where $r=K$, \ie, full source predictions are utilized.
	For comparison, we choose SOTA source-model-based (Mod.) methods (\eg \cite{liang2020we,liang2021source}) and source-data-dependent (Data) methods (\eg, \cite{jin2021re,li2021semantic}) which show the best UDA results so far.
	In addition, we provide the results of a recent ViT-based UDA method---TransDA \cite{yang2021transformer} for fair comparison.
	
	\setlength{\tabcolsep}{3.0pt}
	\begin{table*}[]
		\centering
		\caption{Accuracies (\%) on \textbf{Office-Home} \cite{venkateswara2017deep} for single-source partial-set UDA.}
		\vspace{-5pt}
		\resizebox{0.88\textwidth}{!}{$
			\begin{tabular}{lccccccccccccca}
				\toprule
				Method & Type & Ar$\to$Cl & Ar$\to$Pr & Ar$\to$Re & Cl$\to$Ar & Cl$\to$Pr & Cl$\to$Re & Pr$\to$Ar & Pr$\to$Cl & Pr$\to$Re & Re$\to$Ar & Re$\to$Cl & Re$\to$Pr & Avg. \\
				\midrule
				No Adapt. & Pred. & 44.9 & 70.5 & 80.7 & 57.5 & 61.3 & 67.2 & 60.9 & 40.8 & 76.0 & 70.9 & 47.6 & 76.9 & 62.9 \\
				NLL-OT \cite{asano2019self} & Pred. & 42.7 & 64.2 & 71.7 & 57.2 & 58.5 & 64.5 & 56.7 & 41.6 & 67.5 & 64.2 & 45.1 & 69.0 & 58.6 \\
				NLL-KL \cite{zhang2021unsupervised} & Pred. & 38.9 & 53.8 & 60.5 & 49.2 & 50.5 & 55.9 & 50.0 & 38.9 & 58.0 & 57.0 & 41.7 & 59.6 & 51.2 \\
				HD-SHOT \cite{liang2020we} & Pred. & 51.2 & 76.2 & 85.7 & 68.8 & 70.6 & 77.5 & 69.2 & 49.6 & 81.4 & 75.9 & 54.1 & 80.7 & 70.1 \\
				SD-SHOT \cite{liang2020we} & Pred. & 54.2 & 81.8 & 88.9 & 74.8 & 76.5 & 81.0 & 73.5 & 50.6 & 84.2 & 79.8 & 58.4 & 83.7 & 74.0 \\
				DINE & Pred. & 58.1 & 83.4 & 89.2 & 78.0 & 80.0 & 80.6 & 74.2 & 56.6 & 85.9 & 80.6 & 62.9 & 84.8 & 76.2 \\
				DINE (full) & Pred. & 55.6 & 79.0 & 85.3 & 75.3 & 77.6 & 78.5 & 74.1 & 56.7 & 83.8 & 78.4 & 59.6 & 83.1 & 73.9 \\
				\midrule
				\multicolumn{15}{c}{source backbone: \textbf{ResNet-50 ($\upuparrows$)}, \textbf{ViT ($\downdownarrows$)} $\to$ \textbf{ResNet-50} (target backbone)} \\
				\midrule
				No Adapt. & Pred. & 55.4 & 83.5 & 89.1 & 79.0 & 80.6 & 84.3 & 77.7 & 45.8 & 87.7 & 83.3 & 52.6 & 85.5 & 75.4 \\
				NLL-OT \cite{asano2019self} & Pred. & 49.7 & 75.0 & 80.2 & 68.8 & 73.1 & 77.2 & 69.3 & 44.6 & 80.2 & 75.8 & 49.1 & 78.5 & 68.5 \\
				NLL-KL \cite{zhang2021unsupervised} & Pred. & 45.9 & 59.4 & 62.2 & 57.7 & 59.4 & 59.6 & 56.0 & 41.5 & 60.8 & 58.4 & 44.8 & 60.1 & 55.5 \\
				HD-SHOT \cite{liang2020we} & Pred. & 54.9 & 82.2 & 88.0 & 79.6 & 77.7 & 83.6 & 78.1 & 48.4 & 86.0 & 83.4 & 53.1 & 81.3 & 74.7 \\
				SD-SHOT \cite{liang2020we} & Pred. & 58.2 & 84.1 & 89.7 & 82.5 & 81.7 & 84.8 & 81.8 & 51.0 & 88.9 & 85.8 & 57.7 & 83.6 & 77.5 \\
				DINE & Pred. & \textbf{\color{awesome}67.8} & \textbf{\color{awesome}92.0} & \textbf{\color{awesome}91.8} & \textbf{\color{awesome}84.5} & 89.1 & \textbf{\color{awesome}87.7} & 83.5 & \textbf{\color{awesome}63.9} & \textbf{\color{awesome}91.7} & \textbf{\color{awesome}87.0} & \textbf{\color{awesome}65.9} & \textbf{\color{awesome}91.3} & \textbf{\color{awesome}83.0} \\
				DINE (full) & Pred. & 65.3 & 90.3 & 91.1 & 84.3 & \textbf{\color{awesome}89.7} & 86.4 & \textbf{\color{awesome}83.7} & 62.1 & 91.2 & 86.3 & 64.7 & 90.3 & 82.1 \\
				\midrule
				\midrule
				SHOT \cite{liang2021source} & Mod. & 64.6 & 85.1 & 92.9 & 78.4 & 76.8 & 86.9 & 79.0 & 65.7 & 89.0 & 81.1 & 67.7 & 86.4 & 79.5 \\
				SHOT++ \cite{liang2021source} & Mod. & 66.0 & 86.1 & 92.8 & 77.9 & 77.5 & 87.6 & 78.6 & 66.4 & 89.7 & 81.5 & 67.9 & 87.2 & 79.9 \\
				TransDA* \cite{yang2021transformer} & Mod. & 73.0 & 79.5 & 90.9 & 72.0 & 83.4 & 86.0 & 81.1 & 71.0 & 86.9 & 87.8 & 74.9 & 89.2 & 81.3 \\
				MCC~\cite{jin2020minimum} & Data & 63.1 & 80.8 & 86.0 & 70.8 & 72.1 & 80.1 & 75.0 & 60.8 & 85.9 & 78.6 & 65.2 & 82.8 & 75.1 \\
				JUMBOT~\cite{fatras2021unbalanced} & Data & 62.7 & 77.5 & 84.4 & 76.0 & 73.3 & 80.5 & 74.7 & 60.8 & 85.1 & 80.2 & 66.5 & 83.9 & 75.5 \\
				BA$^3$US~\cite{liang2020balanced} & Data & 60.6 & 83.2 & 88.4 & 71.8 & 72.8 & 83.4 & 75.5 & 61.6 & 86.5 & 79.3 & 62.8 & 86.1 & 76.0 \\
				\bottomrule
			\end{tabular}
			$}
		\label{tab:pda}
		\vspace{-5pt}
	\end{table*}
	
	\subsection{Results}
	\setlength{\tabcolsep}{4.0pt}
	\begin{table*}[!htbp]
		\centering
		\caption{Accuracies (\%) on four different datasets for multi-source closed-set UDA.}
		\vspace{-5pt}
		\resizebox{0.93\textwidth}{!}{$
			\begin{tabular}{llcccacccaccccacccca}
				\toprule
				\multicolumn{2}{c}{Dataset} & \multicolumn{4}{c}{\textbf{Office}~\cite{saenko2010adapting}} & \multicolumn{4}{c}{\textbf{Image-CLEF}~\cite{long2017deep}} & \multicolumn{5}{c}{\textbf{Office-Caltech}~\cite{gong2012geodesic}} & \multicolumn{5}{c}{\textbf{Office-Home}~\cite{venkateswara2017deep}} \\
				\midrule
				Method & Type & $\to$A & $\to$D & $\to$W & Avg.  
				& $\to$C & $\to$I & $\to$P & Avg. 
				& $\to$A & $\to$C & $\to$D & $\to$W & Avg.
				& $\to$Ar & $\to$Cl & $\to$Pr & $\to$Re & Avg. \\
				\midrule
				No Adapt. & Pred. & 64.5 & 82.3 & 80.7 & 75.8 & 92.1 & 87.4 & 72.4 & 84.0 & 84.9 & 88.7 & 93.0 & 88.5 & 88.8 & 54.9 & 49.9 & 69.6 & 76.7 & 62.8 \\
				DINE (w/o FT) & Pred. & 69.2 & 98.6 & 96.9 & 88.3 & 96.2 & 91.4 & 78.3 & 88.6 & 95.0 & 92.0 & 98.5 & 97.3 & 95.7 & 70.8 & 57.1 & 80.9 & 82.1 & 72.7 \\
				DINE & Pred. & 76.8 & \textbf{\color{awesome}99.2} & 98.4 & 91.5 & \textbf{\color{awesome}98.0} & 93.4 & 80.2 & 90.5 & 95.9 & 95.2 & 98.5 & \textbf{\color{awesome}98.9} & 97.1 & 74.8 & 64.1 & 85.0 & 84.6 & 77.1 \\
				DINE (full) & Pred. & 77.1 & \textbf{\color{awesome}99.2} & 98.2 & 91.5 & 97.8 & 93.0 & 79.7 & 90.2 & 96.1 & 95.3 & 98.1 & \textbf{\color{awesome}98.9} & 97.1 & 74.9 & 62.6 & 84.6 & 84.7 & 76.7 \\
				\midrule
				\multicolumn{20}{c}{source backbone: \textbf{ResNet-101 ($\upuparrows$)}, \textbf{ViT ($\downdownarrows$)} $\to$ \textbf{ResNet-101} (target backbone)} \\
				\midrule
				No Adapt. & Pred. & 77.2 & 88.2 & 89.2 & 84.9 & 95.3 & 90.2 & 72.0 & 85.9 & 92.9 & 95.9 & 98.7 & 95.8 & 95.8 & 74.5 & 54.5 & 83.2 & 87.2 & 74.8 \\
				DINE (w/o FT) & Pred. & 80.7 & 98.4 & 97.1 & 92.1 & 97.2 & \textbf{\color{awesome}96.6} & 80.9 & 91.6 & 96.4 & 96.0 & 99.4 & 98.2 & 97.5 & 82.4 & 61.0 & 88.6 & 90.8 & 80.7 \\
				DINE & Pred. & \textbf{\color{awesome}82.4} & \textbf{\color{awesome}99.2} & 98.4 & \textbf{\color{awesome}93.4} & 97.8 & \textbf{\color{awesome}96.6} & 81.3 & \textbf{\color{awesome}91.9} & \textbf{\color{awesome}96.8} & \textbf{\color{awesome}97.0} & 99.6 & 98.8 & \textbf{\color{awesome}98.0} & \textbf{\color{awesome}83.6} & \textbf{\color{awesome}67.0} & \textbf{\color{awesome}90.9} & \textbf{\color{awesome}91.8} & \textbf{\color{awesome}83.3} \\
				DINE (full) & Pred. & 81.4 & 99.0 & \textbf{\color{awesome}98.5} & 93.0 & 97.8 & 96.4 & \textbf{\color{awesome}81.4} & \textbf{\color{awesome}91.9} & 96.7 & 96.4 & \textbf{\color{awesome}99.8} & 98.6 & 97.9 & 83.4 & 65.2 & 90.3 & 91.5 & 82.6 \\		
				\midrule
				\midrule
				SHOT \cite{liang2021source} & Mod. & - & - & - & - 
				& - & - & - & - 
				& 96.2 & 96.2 & 98.5 & 99.8 & 97.7 
				& 73.0 & 60.4 & 83.9 & 83.3 & 75.2 \\
				DECISION~\cite{ahmed2021unsupervised} & Mod. &75.4 & 98.4 & 99.6 & 91.1 
				& - & - & - & - 
				& 95.9 & 95.9 & 100. & 99.6 & 98.0
				& 74.5 & 59.4 & 84.4 & 83.6 & 75.5 \\ 
				SHOT++~\cite{liang2021source} & Mod. & - & - & - & - 
				& - & - & - & - 
				& 96.2 & 96.5 & 99.4 & 100. & 98.0
				& 73.1 & 61.3 & 84.3 & 84.0 & 75.7 \\
				CAiDA~\cite{dong2021confident} & Mod. & 75.8 & 99.8 & 98.9 & 91.6 
				& - & - & - & - 
				& 96.8 & 97.1 & 100. & 99.8 & 98.4
				& 75.2 & 60.5 & 84.7 & 84.2 & 76.2 \\
				SImpAl$_{101}$~\cite{venkat2021your} & Data & 71.2 & 99.4 & 97.9 & 89.5 
				& 95.2 & 91.7 & 78.0 & 88.3 
				& 95.6 & 94.6 & 100. & 100. & 97.5
				& 73.4 & 62.4 & 81.0 & 82.7 & 74.8 \\
				MFSAN~\cite{zhu2019aligning} & Data & 72.7 & 99.5 & 98.5 & 90.2 
				& 95.4 & 93.6 & 79.1 & 89.4 
				& - & - & - & - & -
				& 72.1 & 62.0 & 80.3 & 81.8 & 74.1 \\
				MIAN-$\gamma$ \cite{park2021information} & Data & 76.2 & 99.2 & 98.4 & 91.3 
				& - & - & - & - 
				& - & - & - & - & -
				& 69.9 & 64.2 & 80.9 & 81.5 & 74.1  \\
				PCT~\cite{tanwisuth2021prototype} & Data & - & - & - & - 
				& - & - & - & - 
				& - & - & - & - & -
				& 76.3 & 64.1 & 84.9 & 84.3 & 77.4 \\
				\bottomrule
			\end{tabular}
			$}
		\label{tab:ms}
		\vspace{-10pt}
	\end{table*}
	
	\textbf{a) single-source.} 
	We first show the results of cross-domain object recognition on \textbf{Office} in Table~\ref{tab:office}. and \textbf{Office-Home} in Table~\ref{tab:home}. 
	As stated above, we adopt two different backbone networks for training the source domain. 
	Typically, UDA methods assume the same network structure across domains, \eg, ResNet-50 in \cite{long2018conditional,jin2021re,li2021semantic}.
	With the same ResNet-50 backbone, DINE performs better than other baselines, indicating the effectiveness of the proposed distillation strategy. 
	Trained with much stronger source models like ViT, all the black-box UDA methods are significantly strengthened, and DINE achieves the best mean accuracy 91.9\% on Office and 80.7\% on Office-Home. 
	These results even beat SOTA model-based and data-based UDA methods with clear margins.
	As can be seen from Table~\ref{tab:visda}, the results on \textbf{VisDA-C} again validate the superiority of DINE over other baselines. 
	Similar observations are discovered on this dataset, \ie, the stronger the source model is, the better results black-box UDA methods obtain. 
	Typically, UDA methods always assume the same network structure (ResNet-101) across domains. 
	Compared with these methods via ResNet-101, DINE obtains competitive results and even beats ViT-based TransDA \cite{yang2021transformer}.
	
	\begin{figure*}[!htb]
		\centering
		\small
		\setlength\tabcolsep{1mm}
		\renewcommand\arraystretch{0.1}
		\begin{tabular}{cccc}
			\includegraphics[width=0.24\linewidth,trim={3.3cm 9.2cm 4.2cm 9.8cm}, clip]{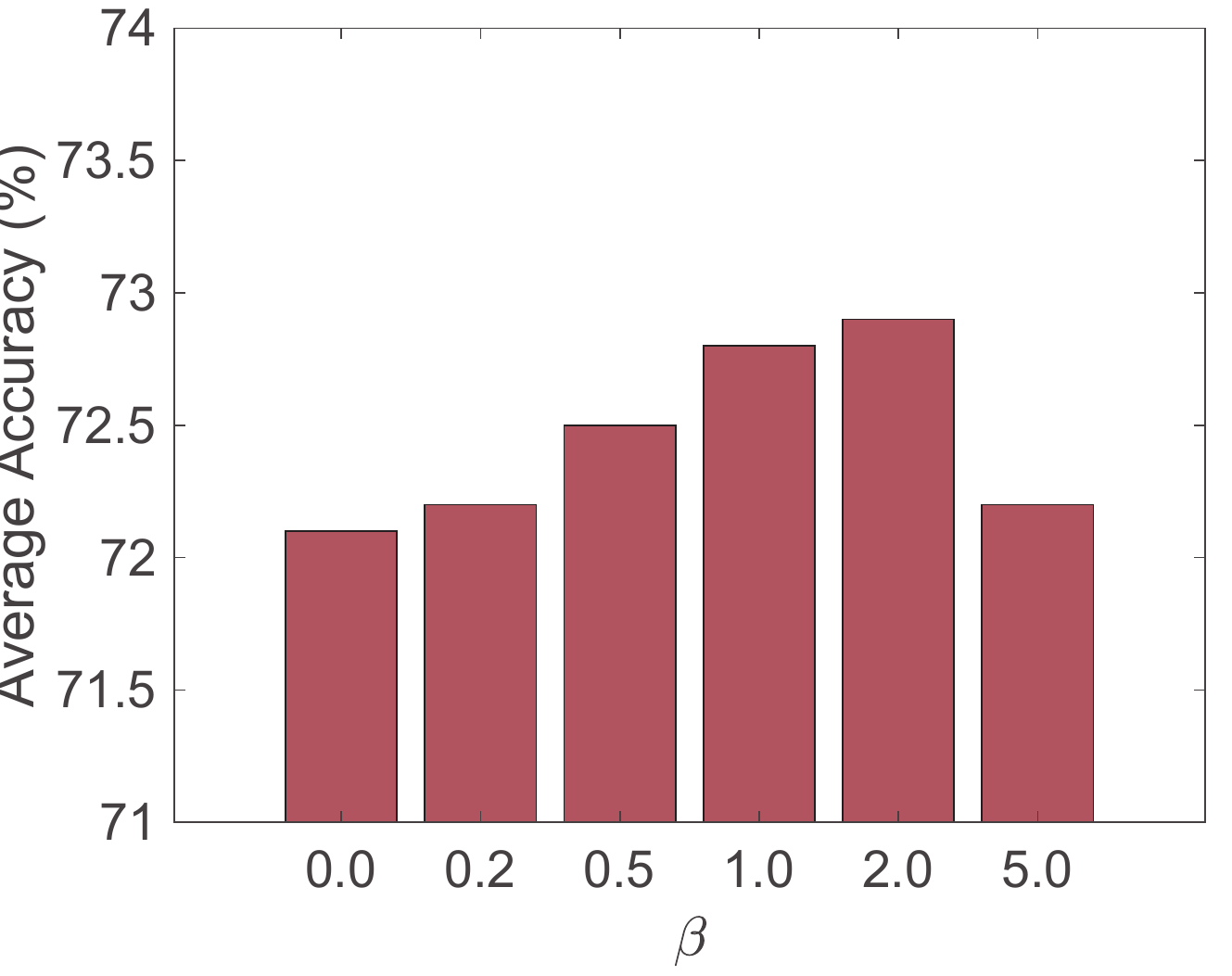} &
			\includegraphics[width=0.24\linewidth,trim={3.3cm 9.2cm 4.2cm 9.8cm}, clip]{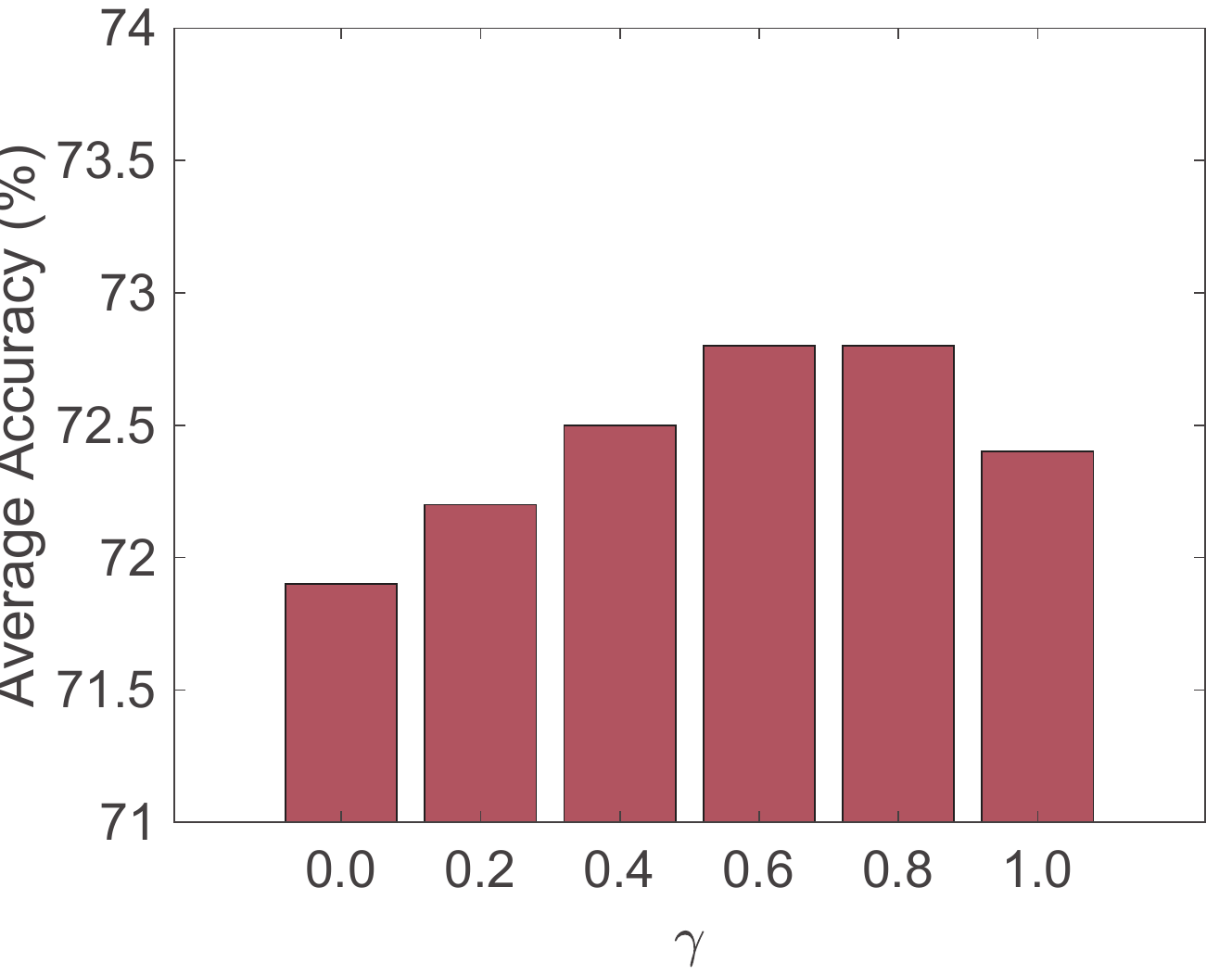} & 
			\includegraphics[width=0.24\linewidth,trim={3.3cm 9.2cm 4.2cm 9.8cm}, clip]{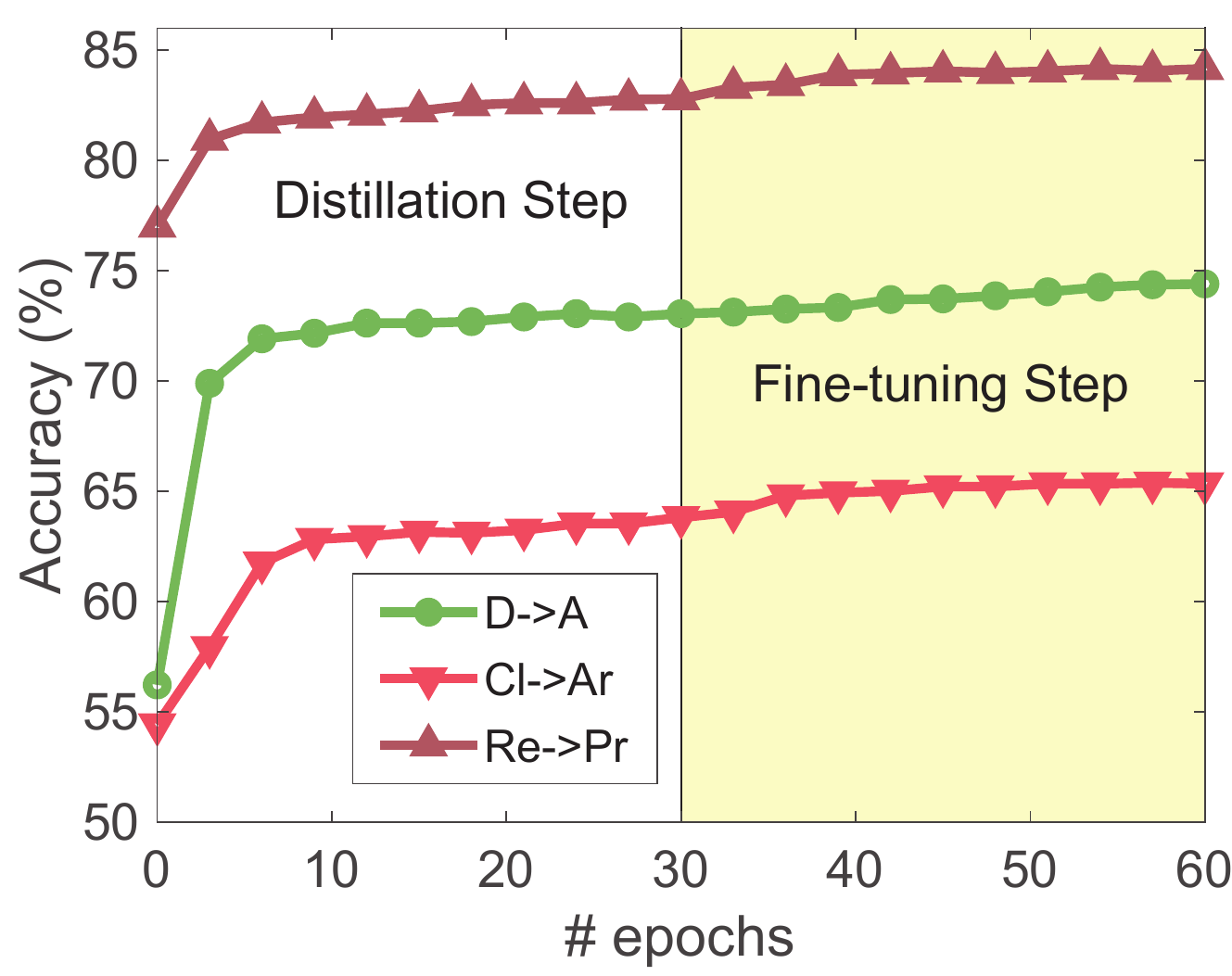} &
			\includegraphics[width=0.24\linewidth,trim={3.3cm 9.2cm 4.2cm 9.8cm}, clip]{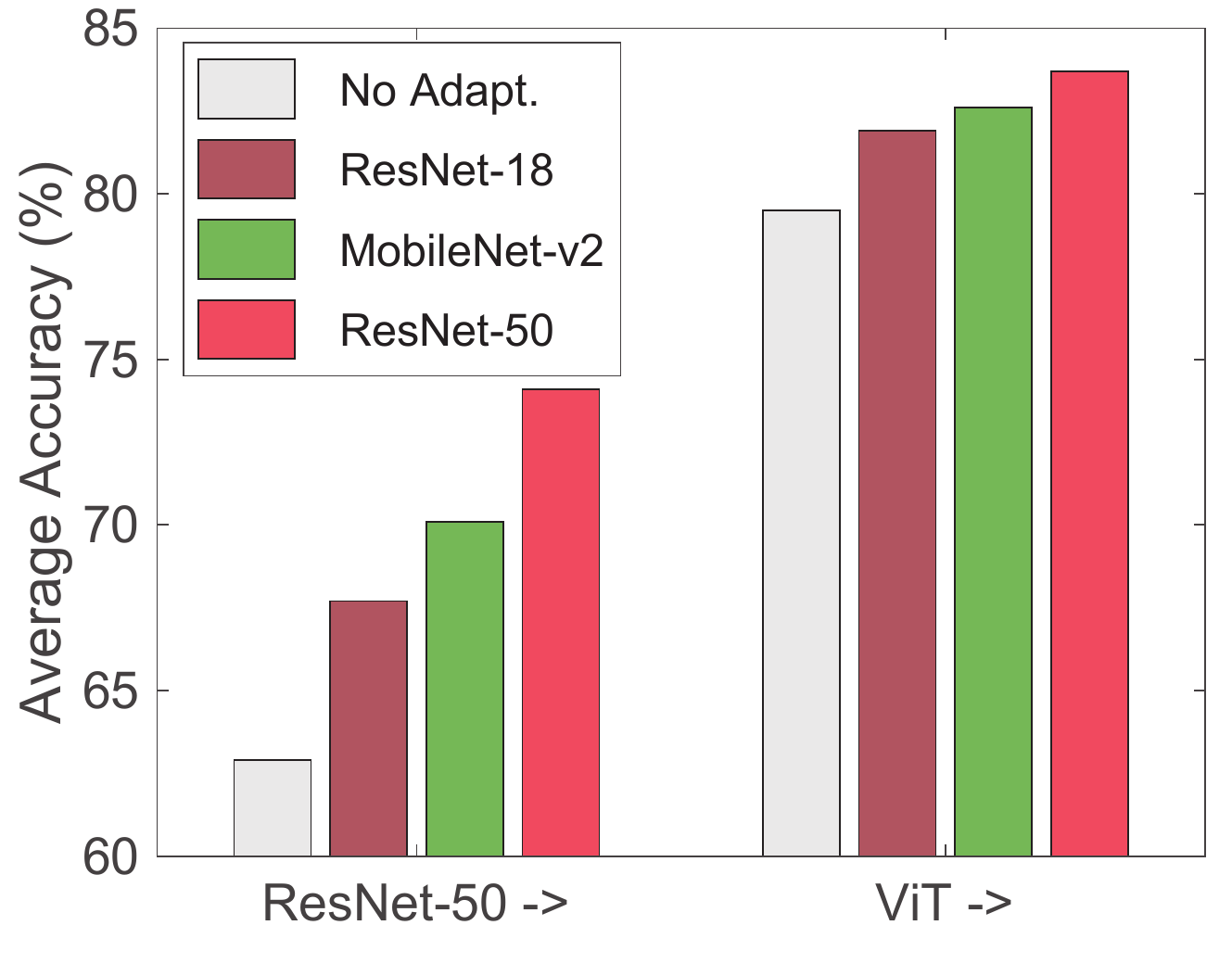} \\
			~\\
			~\\
			(a) sensitivity to $\beta$ ($\gamma=0.6$)  & (b) sensitivity to $\gamma$ ($\beta=1.0$) & (c) accuracy convergence & (d) different target backbones
		\end{tabular}
		\caption{\textbf{Analysis on DINE for three UDA tasks}. (a-b) plot the average accuracies with different values of $\beta,\gamma$; (c) shows the accuracies during the training process; (d) shows the average accuracies when choosing target backbone in \{MobileNet-v2, ResNet-18, ResNet-50\}.}
		\label{fig:par_sen}
		\vspace{-10pt}
	\end{figure*} 
	
	\textbf{b) partial-set.} 
	In addition to the closed-set UDA problem, we also investigate the generalization ability of these black-box methods for partial-set UDA. 
	We follow \cite{cao2018partial,liang2020balanced} and only select the first 25 classes in alphabetical order in the target domain. 
	As shown in Table~\ref{tab:pda}, NLL-OT and NLL-KL even under-perform `No~Adapt.' due to the challenging asymmetric label spaces. 
	Moreover, the proposed DINE achieves consistently better results than HD-SHOT and SD-SHOT.
	Different from previous tables, DINE even outperforms DINE (full) for both source backbones by utilizing the proposed adaptive label smoothing technique.
	With the ViT-based source predictor, DINE obtains the highest average accuracy (83.0\%) which is also fairly higher than previous state-of-the-arts in \cite{liang2020balanced,liang2021source} via ResNet-50 and TransDA.

	\textbf{c) multi-source.} 
	We also study the performance of multi-source UDA \cite{peng2019moment,yang2020curriculum,li2021dynamic} where multiple source domains exist.
	As shown in Table~\ref{tab:ms}, we choose four different multi-source datasets. 
	Within each dataset, we aim to transfer knowledge from the other subsets to the target subset.
	It is found that DINE outperforms `No~Adapt.' and DINE (w/o FT), indicating the effectiveness of fine-tuning and structural distillation.
	Compared with DINE (full), DINE obtains competitive results on all these datasets.
	With the ViT-based source predictor, DINE again beats prior data-based and model-based multi-source UDA methods.

\setlength{\tabcolsep}{4.0pt}
\begin{table}[]
	\centering
	\caption{\textbf{Ablation}. Results of different variants for single-source closed-set UDA with {ResNet-50} (source) $\to$ {ResNet-50} (target).}
	\vspace{-5pt}
	\resizebox{0.44\textwidth}{!}{$
		\begin{tabular}{lccccc}
			\toprule
			Method & FT  & D $\to$ A & Cl $\to$ Ar & Re $\to$ Pr & Avg.\\
			\midrule
			& $\times$ & 62.7$\pm$2.2 & 60.6$\pm$0.7 & 80.4$\pm$0.3 & 67.9 \\
			\multirow{-2}{*}{DINE (w/o $\mathcal{L}_{im}$)} & \cellcolor{ggray}$\checkmark$ & \cellcolor{ggray}63.6$\pm$3.5 & \cellcolor{ggray}59.6$\pm$1.1 & \cellcolor{ggray}81.4$\pm$0.1 & \cellcolor{ggray}68.2 \\
			& $\times$ & 70.2$\pm$1.4 & 63.2$\pm$0.4 & 82.7$\pm$0.4 & 72.1 \\
			\multirow{-2}{*}{DINE (w/o $\mathcal{L}_{mix}$)} & \cellcolor{ggray}$\checkmark$ & \cellcolor{ggray}72.0$\pm$1.7 & \cellcolor{ggray}64.6$\pm$0.2 & \cellcolor{ggray}84.0$\pm$0.3 & \cellcolor{ggray}73.5 \\
			& $\times$ & 70.6$\pm$2.0 & 64.1$\pm$0.3 & 83.7$\pm$0.4 & 72.8 \\
			\multirow{-2}{*}{DINE} & \cellcolor{ggray}$\checkmark$ & \cellcolor{ggray}\textbf{72.2}$\pm$1.8 & \cellcolor{ggray}\textbf{65.3}$\pm$0.2 & \cellcolor{ggray}\textbf{84.7}$\pm$0.1 & \cellcolor{ggray}\textbf{74.1} \\
			\bottomrule
		\end{tabular}
		$}
	\label{tab:ab1}
\end{table}

\setlength{\tabcolsep}{4.0pt}
\begin{table}[]
	\centering
	\caption{\textbf{Study on the AdaLS}. Results for closed-set UDA with {ResNet-50} (source) $\to$ {ResNet-50} (target). Hard: one-hot label.}
	\vspace{-5pt}
	\resizebox{0.44\textwidth}{!}{$
		\begin{tabular}{lccccc}
			\toprule
			Method & FT & D $\to$ A & Cl $\to$ Ar & Re $\to$ Pr & Avg.\\
			\midrule
			& $\times$ & 63.6$\pm$2.1 & 60.9$\pm$0.4 & 80.2$\pm$0.6 & 68.2 \\
			\multirow{-2}{*}{DINE (w/ Hard)} & \cellcolor{ggray}$\checkmark$ & \cellcolor{ggray}67.3$\pm$1.7 & \cellcolor{ggray}62.3$\pm$0.2 & \cellcolor{ggray}81.7$\pm$0.4 & \cellcolor{ggray}70.4 \\
			& $\times$ & 64.8$\pm$1.9 & 61.7$\pm$0.4 & 81.0$\pm$0.5 & 69.2 \\
			\multirow{-2}{*}{DINE (w/ LS)} & \cellcolor{ggray}$\checkmark$ & \cellcolor{ggray}68.4$\pm$1.8 & \cellcolor{ggray}63.0$\pm$0.5 & \cellcolor{ggray}82.6$\pm$0.3 & \cellcolor{ggray}71.3 \\
			& $\times$ & 70.6$\pm$2.0 & 64.1$\pm$0.3 & 83.7$\pm$0.4 & 72.8 \\
			\multirow{-2}{*}{DINE ($r=1$)} & \cellcolor{ggray}$\checkmark$ & \cellcolor{ggray}72.2$\pm$1.8& \cellcolor{ggray}65.3$\pm$0.2 & \cellcolor{ggray}84.7$\pm$0.1 & \cellcolor{ggray}74.1 \\
			& $\times$ & 71.4$\pm$2.5 & 64.4$\pm$0.5 & 83.7$\pm$0.3 & 73.1 \\
			\multirow{-2}{*}{DINE ($r=3$)} & \cellcolor{ggray}$\checkmark$ & \cellcolor{ggray}\textbf{72.9}$\pm$2.0 & \cellcolor{ggray}65.6$\pm$0.5 & \cellcolor{ggray}\textbf{84.7}$\pm$0.3 & \cellcolor{ggray}\textbf{74.4} \\
			& $\times$ & 71.3$\pm$2.7 & 64.6$\pm$0.4 & 83.3$\pm$0.4 & 73.1 \\
			\multirow{-2}{*}{DINE ($r=K$)} & \cellcolor{ggray}$\checkmark$ & \cellcolor{ggray}\textbf{72.9}$\pm$2.5 & \cellcolor{ggray}\textbf{65.9}$\pm$0.7 & \cellcolor{ggray}84.3$\pm$0.3 & \cellcolor{ggray}74.3 \\
			\bottomrule
		\end{tabular}
		$}
	\label{tab:ab2}
	\vspace{-10pt}
\end{table}

	\subsection{Analysis}
	We study the contribution of different components within DINE, with results shown in Table \ref{tab:ab1} and Table \ref{tab:ab2}.
	When $\mathcal{L}_{im}$ or $\mathcal{L}_{mix}$ is dropped, the performance of DINE decrease for all three tasks, verifying their importance.
	The second step called FT is also universally vital, which enhances a variety of variants in two tables.
	Regarding the devised AdaLS technique in Eq.~(\ref{eq:als}), we compare it with one-hot encoding (Hard) and vanilla labeling smoothing (LS).
	With or without the FT step, AdaLS ($r=1$) works better than Hard and LS and is competitive to DINE ($r=K$). 
	If we remain more classes, \eg, $r=3$, AdaLS even works better than using full vectors within DINE.
	
	We study the parameter sensitivity of $\beta, \gamma$ in Fig.~\ref{fig:par_sen}(a-b), where $\beta$ is in the range of [0.0, 0.2, 0.5, 1.0, 2.0, 5.0], and $\gamma$ is in the range of [0.0, 0.2, 0.4, 0.6, 0.8, 1.0].
	It is easy to find the results around the selected parameters $\beta=1.0, \gamma=0.6$ are quite stable.
	Note that other parameters $\beta=2.0, \gamma=0.8$ may be better via oracle validation.
	Besides, in Fig.~\ref{fig:par_sen}(c), in both steps of DINE, the accuracies keep increasing and become convergent. 
	
	\subsection{Discussion \& Limitation}
	To evaluate DINE with small models for a low-resource target user, we provide the results of ResNet-18 and MobileNet-v2 \cite{sandler2018mobilenetv2} being the target backbone, respectively. 
	As shown in Fig.~\ref{fig:par_sen}(d), small target models manage to achieve competitive performance given a strong source predictor like ViT. 
	The main limitation is that this paper does not cover UDA with unknown classes in the target domain \cite{you2019universal,kundu2020universal,saito2020universal}, which needs to be investigated in the future.
	
	\section{Conclusion}
	We explore a new but realistic UDA setting where each source domain only provides its black-box predictor to the target domain, allowing different networks for different domains.
	Thereafter, we propose a simple yet effective two-step structural distillation framework called Distill and Fine-tune (DINE).
	DINE elegantly refines the noisy teacher output via adaptive smoothing and fully considers the data structure in the target domain during distillation.
	Experiments on multiple datasets verify the superiority of DINE over baselines for various UDA tasks.
	Provided with strong pre-trained source models, DINE even achieves state-of-the-art adaptation results with a small efficient target model.

	\textbf{Acknowledgment.}
	This work was partially funded by National Natural Science Foundation of China under Grants U21B2045 and U20A20223; and Beijing Nova Program under Grant Z211100002121108.
	
	\section*{Pseudo code}
	\begin{algorithm}[!htb]
	\caption{Pseudocode of DINE for black-box UDA.}
	\small 
	\setstretch{0.99}
	\label{alg:DINE}
	\begin{algorithmic}
		\STATE {\bfseries 1. Source Model Generation:}
		\STATE {\bfseries Require:} $\{x_s^i,y_s^i\}_{i=1}^{n_s}$, the number of epochs $T_m$.
		\STATE $\triangleright$ Train $f_s$ via minimizing the objective in Eq.~(\ref{eq:ls}).
		\STATE {\bfseries 2. Distillation:} 
		\STATE {\bfseries Require:} Target data $\{x_t^i\}_{i=1}^{n_t}$, $T_m$, parameters $\alpha=0.3,\beta,\gamma$.
		\STATE $\triangleright$ Initialize the teacher output $f_s(x_t)$ via Eq.~(\ref{eq:init}) ($r=1$).
		\FOR {$e=1$ {\bfseries to} $T_m$}
		\FOR{$i=1$ {\bfseries to} $n_b$}
		\STATE $\triangleright$ Sample a batch from target data.
		\STATE $\triangleright$ Apply MixUp within the batch.
		\STATE $\triangleright$ Update $f_t$ via minimizing the objective in Eq.~(\ref{eq:overall}).
		\ENDFOR
		\STATE $\triangleright$ Update the teacher output $f_s(x_t)$ via Eq.~(\ref{eq:ema}).
		\ENDFOR
		\STATE {\bfseries 3. Fine-tuning:}
		\STATE {\bfseries Require:} Target data $\{x_t^i\}_{i=1}^{n_t}$, $T_m$.
		\STATE $\triangleright$ Fine-tune $f_t$ via maximizing the objective in Eq.~(\ref{eq:im}).
	\end{algorithmic}
	\end{algorithm}

	{\small
		\bibliographystyle{ieee_fullname}
		\bibliography{egbib}
	}
	
\end{document}